\documentclass{article}

\usepackage{PRIMEarxiv}

\usepackage[utf8]{inputenc} 
\usepackage[T1]{fontenc}    
\usepackage[hidelinks]{hyperref}       
\usepackage{url}            
\usepackage{booktabs}       
\usepackage{amsfonts}       
\usepackage{nicefrac}       
\usepackage{microtype}      
\usepackage{lipsum}
\usepackage{fancyhdr}       
\usepackage{graphicx}       
\graphicspath{{media/}}     
\usepackage[table,xcdraw]{xcolor}
\usepackage{graphicx}
\usepackage{times}
\usepackage{helvet}
\usepackage{courier}
\usepackage{mathrsfs}
\usepackage{bbold}
\usepackage{algorithm}
\usepackage{algpseudocode}
\usepackage{amsmath,amssymb}
\usepackage{subfig}
\usepackage{url}
\usepackage[round]{natbib}

\DeclareMathOperator{\E}{\mathbb{E}}
\DeclareMathOperator*{\argmax}{arg\,max}
\DeclareMathOperator*{\argmin}{arg\,min}
\newcommand{\vect}[1]{\boldsymbol{#1}}

\title{Maximum entropy exploration in contextual bandits with neural networks and energy based models
}

\author{
  Adam Elwood, Marco Leonardi, Ashraf Mohamed, Alessandro Rozza \\
  lastminute.com group \\
 Vicolo de Calvi, 2, Chiasso, Switzerland \\
}

\begin{document}
\maketitle

\begin{abstract}
    Contextual bandits can solve a huge range of real-world problems. However, current popular algorithms to solve them either rely on linear models, or unreliable uncertainty estimation in non-linear models, which are required to deal with the exploration-exploitation trade-off. Inspired by theories of human cognition, we introduce novel techniques that use maximum entropy exploration, relying on neural networks to find optimal policies in settings with both continuous and discrete action spaces. We present two classes of models, one with neural networks as reward estimators, and the other with energy based models, which model the probability of obtaining an optimal reward given an action. We evaluate the performance of these models in static and dynamic contextual bandit simulation environments. We show that both techniques outperform well-known standard algorithms, where energy based models have the best overall performance. This provides practitioners with new techniques that perform well in static and dynamic settings, and are particularly well suited to non-linear scenarios with continuous action spaces.
\end{abstract}

\section{Introduction}

In recent years, machine learning has been applied to solve a large array of concrete scientific and business problems \citep{Silver,portugal2018use,Sarker}. The rapid advancements have mainly been due to the increased access to large datasets and computing resources. However, many real world scenarios require online decision making. They generally do not come with readily available datasets that cover the phase space in question, instead the data must be collected as decisions are made. These kind of problems generically come under the banner of reinforcement learning, where a sequential series of actions must be made in an environment, where previous decisions influence future decisions. 

One class of reinforcement learning problem that is particularly relevant to modern technology businesses is known as the contextual bandit, an extension of the multi-armed bandit problem \citep{bouneffouf2020survey}. In contextual bandit algorithms, actions must be chosen given the state of the system, which is specified by its context. Actions are chosen so as to maximise the total reward over time. The result of a particular action is obtained immediately and can be used to inform future decisions. For optimal performance, these actions should be chosen to trade-off exploration of phase space with exploitation of the most rewarding behaviour. Contextual bandits are relevant in many business applications, such as dynamic pricing and recommender systems.

There are lots of machine learning models capable of making predictions about a reward given an input action and context. Artificial neural networks (NNs) are one of the most popular choices. However, these models are typically brittle, in that they still give confident answers outside of the data distribution they have been trained on, where they are likely to be wrong. A policy for choosing actions in a contextual bandit scenario therefore needs an exploration component added on top of the underlying reward estimator. 

One approach to the above issue is to estimate the uncertainties in the predictions made by the neural network. Actions can then be chosen via Thompson sampling, a bayesian methodology for making sequential decisions under uncertainty \citep{thompson1933likelihood,agrawal2013thompson,pmlr-v32-gopalan14}. However, finding accurate and efficient ways of estimating the uncertainties for remains challenging. 

Another approach is maximum entropy exploration, sometimes known as \emph{Active Inference} or \emph{Boltzmann exploration}. This is also popular in Neuroscience as a model of the way the human brain works \citep{FRISTON200670,Friston2009,Friston2010,Brown2012,Adams2013,Schwartenbeck2013,markovic2021empirical,smith2022step}. In maximum entropy exploration, a policy is built that maintains a high entropy over the action space, ensuring it tries lots of different actions, while still aiming for the best possible reward. This has been introduced for contextual bandit problems with a discrete action space \citep{lee2020no}. In this work we extend this approach to work with a continuous action space. 

Energy Based Models (EBMs) are particularly well suited to maximum entropy exploration, due to the close relationship of EBMs with Boltzmann distributions \citep{levine2018reinforcement}. While straightforward neural networks trained with cross-entropy or mean-squared-error losses can work well as reward estimators, they are prone to brittleness. Conversely, EBMs naturally build uncertainty into their formalisation. Instead of giving a certain answer on the best action to play, energy based functions give a degree of possible actions based on the shape of the energy function. Actions can then be found by sampling from this function with techniques based on Markov Chain Monte Carlo (MCMC). These types of models have been considered in full reinforcement learning scenarios \citep{haarnoja2017reinforcement,Du2019ModelBP}. In this work, we introduce a method to apply EBMs based on NNs to contextual bandit problems. 

In this paper we introduce two new contextual bandit algorithms based on maximum entropy exploration. Both algorithms are able to make decisions in continuous action spaces, a key use case that has not been studied as thoroughly as discrete action spaces. Our main contributions can be summarised as follows:

\begin{itemize}
    \item Introducing a technique for maximum entropy exploration with neural networks estimating rewards in contextual bandits with a continuous action space, sampling using Hamiltonian Monte Carlo;
    \item A novel algorithm that uses Energy Based Models based on neural networks to solve contextual bandit problems;
    \item Testing our algorithms in different simulation environments (including with dynamic environments), giving practitioners a guide to which algorithms to use in different scenarios.
\end{itemize}

\section{RELATED WORK}

As they are very relevant to many industry applications, contextual bandits have been widely studied, with many different algorithms proposed, see for example \citep{bietti2021contextual} and \citep{bouneffouf2020survey}. 

Many of the most successful algorithms rely on linear methods for interpreting the context, where it is easier to evaluate output uncertainty \citep{abbasi2011improved,agrawal2013thompson}. This is necessary, because the most commonly applied exploration strategies, Thompson Sampling \citep{thompson1933likelihood} and the Upper Confidence Bound (UCB) algorithm \citep{lai1985asymptotically}, rely on keeping track of uncertainties and updating them as data are collected. However, several techniques for non-linear contextual bandit algorithms have been proposed, using methods based on neural networks with different approaches to predict uncertainties in the output \citep{riquelme2018deep,zhou2020neural,zhang2020neural,kassraie2022neural}.

\subsection{Entropy based exploration in contextual bandits}

As an alternative to Thompson Sampling and UCB, in this work we focus on entropy based exploration strategies, with an emphasis on their application to non-linear contextual bandit problems. This approach has been researched in the reinforcement learning  \citep{kaelbling1996reinforcement,sutton2018reinforcement} and Multi Armed Bandit literature \citep{kuleshov2014algorithms,markovic2021empirical}. 

For the contextual bandit use case, non-linear maximum entropy based exploration with a discrete action space has been considered by \citep{lee2020no}. In this case the non-linearity comes from neural networks, which are used to estimate a reward.

\subsection{Energy based models in reinforcement learning}

Many problems in machine learning, contextual bandits included, revolve around modelling a probability density function, $p(\vect{x})$ for $\vect{x}\in\mathbb{R}^D$. These probability densities can always be expressed in the form of a scalar energy function $E_\theta(\vect{x})$ \citep{lecun2006tutorial}:
\begin{equation}
\label{eq:ebm_basic}
 p(\vect{x}) = \frac{\exp(-E_\theta(\vect{x}))}{\int_{\vect{x}’} \exp(-E_\theta(\vect{x}’)) d\vect{x}’}
\end{equation}
which allows many machine learning problems to be reformulated as an energy based modelling task \citep{grathwohl2019your}. The difficulty with this reformulation comes in estimating the integral in the denominator of Eq.~\ref{eq:ebm_basic}, which is usually intractable. However, if this difficulty can be overcome, a scalar function, $E_\theta(\vect{x})$, is learned which can be evaluated at any value of $\vect{x}$, providing a fully generative model.

Another advantage of EBMs in a reinforcement learning setting is that sampling from them naturally leads to maximum entropy exploration \citep{levine2018reinforcement} . This has been applied to solve full reinforcement learning problems both in model-based \citep{Du2019ModelBP} and a model-free \citep{pmlr-v24-heess12a,haarnoja2017reinforcement} formulations. However, it has not yet been applied to specifically solve the contextual bandit problem.

\section{ALGORITHMS TO SOLVE CONTEXTUAL BANDIT PROBLEMS WITH MAXIMUM ENTROPY EXPLORATION}

In this section we introduce two classes of algorithms for solving contextual bandit problems with NNs, using exploration strategies based on entropy maximisation. In each case the algorithm defines a policy, $\pi(a|\vect{s}_i)$, which gives the probability of playing action $a$, given the observation of state $\vect{s}_i$. The policy is applied by sampling actions from the policy, $a\sim \pi$, at a particular time step. This policy is then updated given the rewards observed in previous time steps by retraining the NNs. 

\subsection{Contextual bandit problem formulation}

Contextual bandit problems require an algorithm to make the choice of an action, $a \in \mathcal{A}$ (where $\mathcal{A}$ is an action space) upon observing the context state of the environment, $s \in \mathcal{S}$ (where $\mathcal{S}$ is a context space). Upon making the action, a reward, $r \in \mathcal{R}$, is received. For each state observed, an action is chosen and the reward recorded. This results in a dataset, $\mathcal{X}$, being built up over a run, consisting of triplets, $\{\vect{s}_i,a_i,r_i\}\in{\mathcal{X}}$, where $i\in\mathbb{N}$ is the time step of a particular triplet. At any step $i$, the data available for choosing the action $a_i$ consists of the set of triplets $\{\vect{s}_j,a_j,r_j\}$ where $j<i$.

The goal of the problem is to maximise the expected reward over an indefinite time horizon, where an arbitrary number of actions, $N$, can be played. This is usually measured in terms of the regret:
\begin{equation}
    \label{eq:regret}
    \mathscr{R}_N = \sum_{i=0}^N [r^*_i - r^{a_i}_i]
\end{equation}
where $r^*_i$ is the best possible reward at the time step $i$ and $r^{a_i}_i$ is the reward at time step $i$ received by the action played, $a_i$. A more successful action choosing policy will have a lower regret.

In this work we assume $\mathcal{A}\in\mathbb{R}$; $\mathcal{S}\in\mathbb{R}^n$, where $n$ is the dimension of the context vector and depends on the particular problem being considered; and $\mathcal{R} \in \mathbb{R}$, where many of the problems considered assume $\mathcal{R} \in [0,1]$.

\subsection{Maximum entropy exploration}

First we define a reward estimator, $\hat{r}_{\theta}(\vect{s}_i,a)$, which gives the expected reward of action $a$ in state $\vect{s}_i$ and is parameterised by the vector $\theta$, similar to the approach in \citep{lee2020no}. In maximum entropy exploration, the policy is defined as follows:
\begin{equation}
\label{eq:max_ent}
    \pi(a | \vect{s}_i) = \argmax_\pi(\E_{a\sim\pi}[\hat{r}_{\theta}(\vect{s}_i,a)] + \alpha \mathcal{H}(\pi))
\end{equation}
where $\mathcal{H(\pi)} = \E_{a\sim\pi}[-\log (\pi)]$ is the Shannon entropy. This can then be solved with a softmax:
\begin{equation}
\label{eq:softmax}
    \pi(a | \vect{s}_i) = \frac{e^{\hat{r}_\theta(a,\vect{s}_i)/\alpha}}{\int_{a'} e^{\hat{r}_\theta(a',\vect{s}_i)/\alpha} da'}
\end{equation}

This approach finds a policy that trades off maximising the expected reward with a chosen action (the first term in Eq.~\ref{eq:max_ent}) with trying a range of different actions, which give a large Shannon entropy (the second term in Eq.~\ref{eq:max_ent}). The degree of this trade off is controlled by the $\alpha \in \mathbb{R}^+$ parameter, which is typically chosen to be at the same scale as the expected reward. Larger $\alpha$ values result in more exploration. Models for $\hat{r}_\theta$ should be chosen to have a fairly flat prior across the states and actions upon initialisation, which will encourage exploration in the early stages of a contextual bandit run. As time progresses and $\hat{r}_\theta$ becomes more certain, the entropy term ensures exploration isn't reduced too prematurely. In the case of static environments, it is also desirable to reduce the $\alpha$ value over time, ensuring the total regret of the algorithm is bounded as $N\to\infty$ \citep{cesa2017boltzmann,lee2020no}.

\subsection{Maximum entropy exploration with neural networks modelling reward}
\label{sec:ann}

We build our reward estimator, $\hat{r}_\theta(a,\vect{s}_i)$, with a neural network trained to predict the reward, $r_i$. With this methodology, we model the expected reward given a certain action. This then allows us to select an action based on this expectation value. As there is no explicit model of the environment, this can be thought of as analogous to the suite of ``model free'' techniques in reinforcement learning \citep{degris2012model}.

In the general case, the neural network can be treated as a regressor with a loss based on the mean-squared-error. In the binary reward case ($\mathcal{R} \in [0,1]$), the network can be treated as a classifier and trained with the binary cross-entropy. Given the reward estimator, samples are drawn from $\pi$ to choose actions online, and $\hat{r}_\theta$ is refit as we collect more data. However, due to the fact that the integral in the denominator of Eq.~\ref{eq:softmax} is likely intractable, sampling is not always trivial. We therefore take two different approaches to approximate the integral.

\subsubsection{Discrete action sampling.}

In the case that there are a predefined discrete set of actions, Eq.~\ref{eq:softmax}, can be rewritten as a sum over all possible actions, $a'$, and explicitly calculated:

\begin{equation}
\label{eq:softmax}
    \pi(a | \vect{s}_i) = \frac{e^{\hat{r}_\theta(a,\vect{s}_i)/\alpha}}{\sum_{a'} e^{\hat{r}_\theta(a',\vect{s}_i)/\alpha}}
\end{equation}

This has the advantage of being easy to implement and apply to a wide range of contextual bandit problems. However, it has the limitation that the time for calculating $\pi$ scales linearly with the number of actions to be sampled, so is not easily applicable to problems with large or continuous action spaces.

\subsubsection{Continuous action sampling.}

Equation~\ref{eq:softmax} has the form of a posterior probability distribution, which is well known in Bayesian statistics, so techniques for sampling from this distribution are widely covered in the literature. To draw samples from $\pi(a|\vect{s}_i)$ in a continuous action space, we can make use of MCMC sampling algorithms. In our case we employ the Hamiltonian Monte Carlo (HMC) algorithm \citep{neal2011mcmc,betancourt2015hamiltonian}, due to its wide usage and availability of suitable implementations \citep{10.1214/aos/1018031103,dillon2017tensorflow}.

This solution works in the general case of a continuous action space, where $a\in\mathbb{R}$. However, in many cases the action space is constrained, such that any particular action $a$ is subject to $a\in [a^{lower},a^{upper}]$, where $a^{lower}$ and $a^{upper}$ denote the upper and lower bounds of possible actions. To deal with this constraint, we can modify the reward estimator to include the constraints, $r^c_\theta$ to return a large negative number when the action is outside the bounds:
\begin{equation}
\begin{split}
    \hat{r}^c_\theta(a,\vect{s}_i) &= \hat{r}_\theta(a,\vect{s}_i) ~\text{when}~ a\in [a^{lower},a^{upper}]\\
    \hat{r}^c_\theta(a,\vect{s}_i) &= -\infty~~~~~~~~ \text{otherwise}.
    \end{split}
\end{equation}
Replacing $\hat{r}_\theta$ with $\hat{r}^c_\theta$ in Eq.~\ref{eq:softmax} when carrying out the HMC sampling will then ensure the actions that are sampled are within the constraints.

A summary of the algorithms described in this section can be seen in Algorithm~\ref{alg:nn}, where the sampling procedure will change depending on whether the action space is discrete or continuous, as described above. In the following, these algorithms will be named \emph{NN Discrete} or \emph{NN HMC} respectively.

\begin{algorithm}
\caption{Maximum entropy exploration with neural networks. \label{alg:nn}}
\begin{algorithmic}
\State Input: $\alpha, N, \theta_0, \mathcal{X}_0, k$

\For{$i=1, … , N$}

\State Receive context $\vect{s}_i$ and choose $a_i\sim \pi_i$ where
\State $\pi_i(a | \vect{s}_i) = \frac{e^{\hat{r}_{\theta_{i-1}}(a,\vect{s}_i)/\alpha}}{\int_{a'} e^{\hat{r}_{\theta_{i-1}}(a’,\vect{s}_i)/\alpha} da'}$

\State Agent receives reward $r_i$
\State Add the triplet $\{ \vect{s}_i, a_i, r_i \}$ to the dataset $\mathcal{X}$
\State Every $k$ steps train the model $\hat{r}_\theta$:
\State $\theta_i = \argmin_\theta \sum_{\{ s_j, a_j, r_j \} \in X} | r_j - \hat{r}_\theta(a_j, s_j) |$

\EndFor 
\end{algorithmic}
\end{algorithm}

\subsection{Maximum entropy exploration with energy based models (EBMs)}

Energy based models allow us to model the probability of choosing an action given a reward, $p(a|\vect{s}_i,r)$, with a scalar-valued energy function $E_\theta(a, \vect{s}_i, r)$, parameterised by $\theta$, which is then marginalised over the state and reward spaces:
\begin{equation}
    p(a|\vect{s}_i,r) = \frac{\exp(-E_\theta(a, \vect{s}_i, r))}{\int_{a'} \exp(-E_\theta(a', \vect{s}_i, r)) da'}
\end{equation}

This then allows us to find the optimal policy by finding the probability of an action to obtain an optimal reward, $r^*$: 

\begin{equation}
    \pi(a|\vect{s}_i) = \frac{\exp(-E_\theta(a, \vect{s}_i, r^*)/\alpha)}{\int_{a'} \exp(-E_\theta(a', \vect{s}_i, r^*)/\alpha) da'}
\end{equation}

Drawing samples of actions from this distribution with MCMC techniques will naturally carry out a maximum entropy exploration policy, where the degree of exploration can again be controlled by the size of $\alpha$ \citep{levine2018reinforcement}.

This methodology is particularly well suited to the case of binary rewards, $\mathcal{R} \in [0,1]$, as it is easy to choose the optimal reward: $r^* =1$.

Contrary to the previous algorithm, when solving the contextual bandit problem with EBMs, we model the probability of an action acting on the environment so as to obtain a certain reward. This is analogous to modelling the probability of a state transition on the environment, so is more in line with the ``model based'' techniques discussed in the reinforcement learning literature \citep{moerland2020model,kaiser2019model,Yilun2019}.

\subsubsection{Training EBMs to solve contextual bandit problems.}

A generic energy function can be learned by minimising $E_\theta(a, \vect{s}, r)$ for the most probable $\{\vect{s},a,r\}$ triplets and maximising it for the least probable triplets that currently have a low energy \citep{lecun2006tutorial}. A simple form of loss function that achieves this goal is \citep{boney2020regularizing,du2019implicit}:
\begin{equation}
    \mathcal{L} = \mathbb{E}_{\emph{x}^+\sim p_D}(E_\theta(\emph{x}^+)) - \mathbb{E}_{\emph{x}^-\sim p_\theta}(E_\theta(\emph{x}^-)),
\end{equation}
where $\emph{x}^+$ represent $\{\vect{s},a,r\}$ triplets drawn from the historical dataset, $\mathcal{X}$, while $\emph{x}^-$ represent triplets sampled from the model.

This approach works in the general case and has the advantage of learning a generative model, which can be used to find any conditional probability distribution. However, training in this way is intensive and unstable, as it requires MCMC sampling when evaluating the loss function and a large existing dataset. 

In the contextual bandit use case, however, we are only interested in learning $\pi(a|\vect{s}_i)$ for the optimal reward, so we can simplify this approach by reducing the input dimensions to the energy function and only learning energies for the optimal rewards:
\begin{equation}
    E_\theta(a, \vect{s}_i)\equiv E_\theta(a, \vect{s}_i, r^*).
\end{equation}
This approach is both easier to train and requires fewer initial training examples. 

After experimenting with the different forms for $\mathcal{L}$ described in \citep{lecun2006tutorial}, we settled on a logarithmic form, which had the most consistent stable performance:
\begin{equation}
    \mathcal{L} = \log(1+ \exp(E_\theta(a^+,\vect{s})-E_\theta(a^-,\vect{s}))),
\end{equation}
where $a^+$ are actions that result in an optimal reward, $r^*$, and  $a^-$ are actions that result in a suboptimal reward (0 in the binary case). 
These values and their corresponding states, $\vect{s}_i$, are taken from historical data.

\subsubsection{Architectures for EBMs.}

To be able to train EBMs it is convenient to choose an architecture that can easily be updated with stochastic gradient descent, while avoiding instabilities in the training. Such instabilities include arbitrarily large or small energy values and energy collapse, where a model learns a minimal value of the energy function across all input values. These criteria can be fulfilled by combining two neural networks, $f_\phi$ and $g_\psi$,  in an \emph{implicit regression} architecture \citep{lecun2006tutorial}, where:
\begin{equation}
\begin{split}
    f_\phi \colon a \to \mathbb{R}\\
    g_\psi \colon \vect{s}_i \to \mathbb{R}
\end{split}
\end{equation}

The energy function can then be defined in two different ways, either linear:
\begin{equation}
    E(a,\vect{s}_i) = |f_\phi(a) - g_\psi(\vect{s}_i)|
\end{equation}
or quadratic:
\begin{equation}
    E(a,\vect{s}_i) = \frac{1}{2}(f_\phi(a) - g_\psi(\vect{s}_i))^2.
\end{equation}
In both these cases, the energy function is bounded from below by 0 and requires two independent networks to both learn the same value for all inputs to result in energy collapse. Both of these features combine to help improve training stability. In the following we consider the quadratic combination.

The added advantage of using neural networks in the architecture is that it allows us to easily draw MCMC samples from $\pi$ using Stochastic Gradient Langevin Dynamics (SGLD), as presented by \citep{du2019implicit} and \citep{song2019generative}. This algorithm works by starting from a random point, $\tilde{x}^0$, and iterating in the direction of higher probability with the gradients of the energy function \citep{debmtut}. Noise, $\omega \sim \mathcal{N}(0,\sigma) $, is added to each gradient step to ensure that the sampling fully captures the underlying probability distribution. This chain is carried out for $K$ steps, where the $k$-th step, $\tilde{x}^k$, is calculated as follows:
\begin{equation}
    \tilde{x}^k \leftarrow \tilde{x}^{k-1} - \eta  \nabla_x E_\theta(\tilde{x}^{k-1}) + \omega
\end{equation}
where $\eta$ is the sample gradient step size.

The full procedure required to solve the contextual bandit problem with energy based models is summarised in Algorithm~\ref{alg:ebm}. In the rest of this work this algorithm will be referred to as the \emph{EBM} algorithm.

\begin{algorithm}
\caption{Contextual bandit with Energy Based Models}\label{alg:ebm}
\begin{algorithmic}
\State Input: $N, \theta_0, \mathcal{X}_0, K, c, \alpha,  a_{max}, a_{min}, \eta, \sigma$

\For{$i=1, … , N$}

\State Choose $a_i\sim \pi_i$ with SGLD, $\tilde{a}^0 \sim U(a_{min},a_{max})$ 

\For{$k=1, … , K$}

\State Draw sample for noise $\omega \sim \mathcal{N}(0,\sigma)$

\State $\tilde{a}^k \leftarrow \tilde{a}^{k-1} - \eta  \nabla_x E_{\theta_{i-1}}(\tilde{a}^{k-1}, \vect{s}_i )/\alpha + \omega$

\EndFor

\State Play action $\tilde{a}^K$, receive $r_i$, update $\mathcal{X}$
\State Every $c$ steps train $E_\theta$ in batches:

\State $\theta_i = \argmin_\theta \sum_{X} \log(1+ e^{E_\theta(a^+,\vect{s}_j)-E_\theta(a^-,\vect{s}_j)})$

\EndFor 
\end{algorithmic}
\end{algorithm}

\subsubsection{Evolution of the energy distribution with dataset size.}

One key property of any model for solving contextual bandit problems is that its uncertainty about the correct action to play decreases as more relevant data is collected. This ensures convergence on the best strategy, which gradually reduces the exploration over time as the space of plausible actions decreases.

\begin{figure*}[ht]
    \centering
    \includegraphics[width=\linewidth]{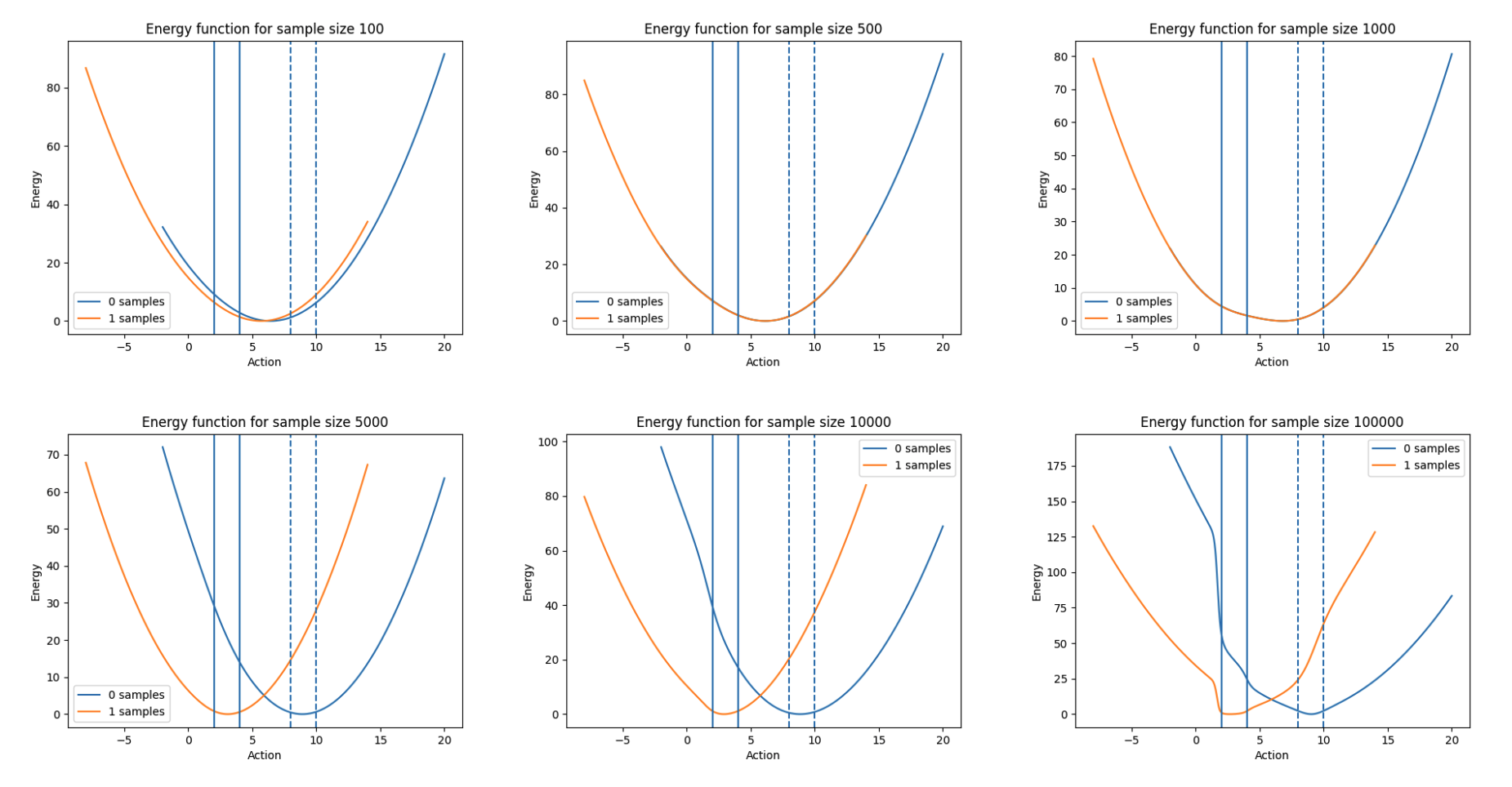}
    \caption{The evolution of an energy function as the training sample size is increased from 100 to 100~000 in an environment with two distinct context categories (labeled 0, blue, and 1, orange) with optimal actions around 3 and 9 respectively.}
    \label{fig:energy_evo}
\end{figure*}

For an energy based model, this is visible as the energy function decreasing in width around the optimal action ranges as the number of samples used for training increases. With the model described in this section, we have empirically justified that we obtain this desired behaviour. An example of this can be seen in Fig.~\ref{fig:energy_evo}, where the evolution of the energy function is plotted as the number of samples are increased for a training set with two contexts, which both have distinct optimal action ranges. It can be seen that as the number of samples increases the model learns to distinguish the two different contexts and narrows in on the optimal action range. Any appropriate sampling approach will therefore sample widely initially and then converge onto the optimal action for each context.

\section{EXPERIMENTS}

To test the algorithms presented so far, we carried out a series of experiments in different contextual bandit simulation environments. We consider both static environments, where the optimal action given a context does not vary over time, and dynamic environments, where it does. This is of particular relevance to algorithms that will be deployed in real-world environments, which are almost never static. This fact also motivated us to focus on settings with continuous action spaces, which are particularly relevant to many industrial use cases. We focus on experiments in a simulation environment due to the lack of relevant benchmarks with dynamic and continuous action spaces in the literature with real datasets.

\subsection{The simulation environments}
    Our simulation environment requires a model to play a series of actions, so as to maximise its total expected reward. Any particular reward is obtained immediately after an action. Let $N$ be the number of rounds, $\vect{s} \in \mathbb{R}^h$ the context vector that the policy observes, and $r_i \in [0,1]$ the the reward given by playing the action $a_i \in \mathbb{R}$ on round $i$ given the context $\vect{s}_i$.
    
    Let $J$ be the number different reward functions $\rho_j \colon a \to [0,1]$. Each context, $\vect{s}$, belongs to a particular reward function, $\rho_j$.
    
    The reward function $\rho_j$ is modelled by the probability density function of a Gaussian distribution $\mathcal{N}(\mu_j,\sigma^2_j)$ parameterized by $\mu_j$ and $\sigma^2_j$, where $\mu_j$ indicates the optimal action that has to be played for that particular reward function.
    Given an action, $a$, the reward function first computes the expected probability of having a reward of 1 by taking the value of the gaussian at $a$, $P_j(r=1|a)$. It then draws a sample from an uniform distribution between 0 and 1, $u_a\sim U(0,1)$ and uses it to calculate the reward:
    \begin{equation}
        \rho_j(a)=\left\{\begin{matrix} 1 & \text{if}~ u_a < P_j(r=1|a) \\  0 & \text{otherwise} \end{matrix}\right.
    \end{equation}
    An example of reward functions for a two context environment can be seen on the left of Fig.~\ref{fig:environment}.    
    
    To make the environment dynamic, it is possible to just modify $\mu_j$ based on the current round of the simulation $i$, making it a function of the timestep, $\mu_j(i)$.
    
    \subsection{Experimental setup}
        We have designed our synthetic environment to have multiple homogeneous contexts, each of which is associated to a reward function. 
    
\begin{figure*}[ht]
    \centering
    \includegraphics[width=\linewidth]{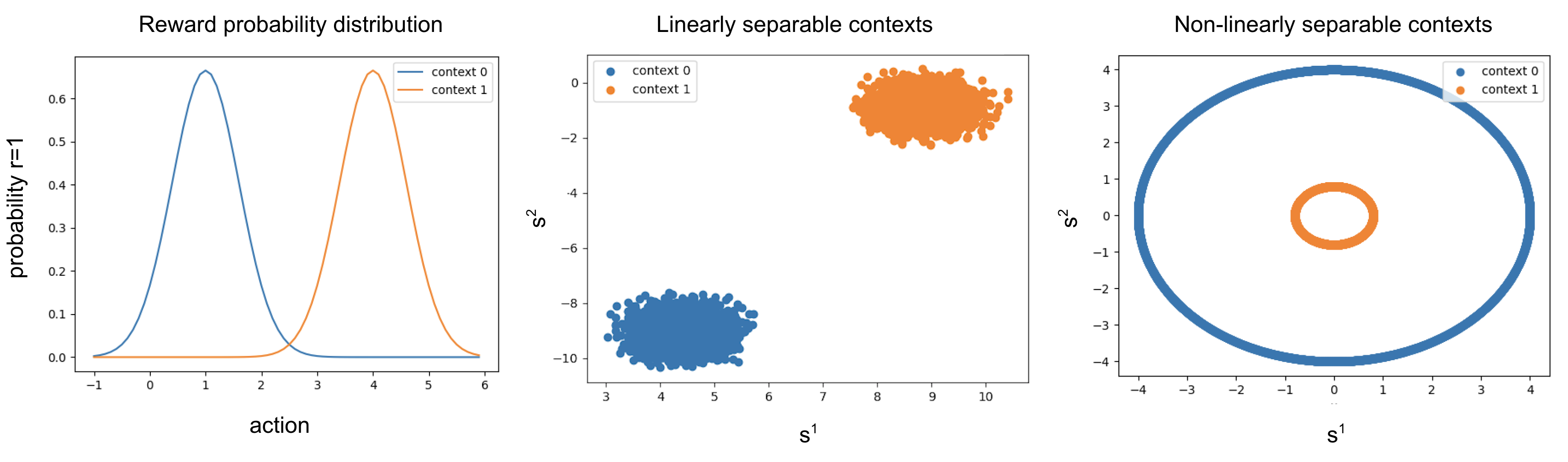}
    \caption{Example of a simulation environment used for testing the algorithms with two contexts. The probability of receiving a reward given an action depends on the context as shown on the left. The linearly and non-linearly separable contexts are shown in the centre and on the right respectively. }
    \label{fig:environment}
\end{figure*}

        To test the capabilities of each proposed method, contexts, $\vect{s}$, can be either linearly, or non-linearly, separable. 
        For the linear case, we generate multiple isotropic Gaussian blobs in a three-dimensional space, $h=3$. Each blob is generated from a Gaussian with fixed standard deviation of 0.4 and a random mean. 
        In the non-linearly separable case, two different set of contexts are generated in a two-dimensional space, $h=2$: a large circle containing a smaller one. Both the circles are zero centered, and have a radius of 4.0 and 0.8 respectively. Examples of similar contexts can be seen on the right and in the centre of Fig.~\ref{fig:environment}.
    
        Experiments consists of $N = 10~000$ observations with $J=2$ different reward functions. They both have the same variance 0.6. In the static environments setting, the mean value, $\mu_j$, is set to 1 and 4 respectively, while in the dynamic setting, it is perturbed by a cosine function:
        \begin{equation}
            \mu_j(i) = \mu_j + \cos\left(\frac{i}{500}\right) + 0.5
        \end{equation}

\subsection{Baseline algorithms}

     We consider the most popular bandit algorithms as baselines against which we compare our algorithms. As a baseline that doesn't take into account the context, we look at two Multi–Armed Bandit (MAB) approaches, namely the Upper Confidence Bound bandit algorithm (UCB1) \citep{lai1985asymptotically} and Thompson Sampling (TS) \citep{thompson1933likelihood}. These algorithms have an arm for each action and learn the best action to play on average. As a baseline that takes into account the context, we have chosen the linear UCB algorithm (linUCB) \citep{abbasi2011improved} and the linear TS algorithm (linTS) \citep{agrawal2013thompson}.

\subsection{Specific configurations of the algorithms}
    Before running all the experiments, all the parameters of the methods were manually fine tuned to achieve the best results.
        
    For the discrete action algorithms (including the MAB and linUCB algorithms), the possible actions are chosen to be between 0.2 and 5.2 with an offset of 1, giving 6 possible actions. None of the possible discrete actions are set on the optimal action point for the simulation environment to avoid giving the discrete algorithms an unfair advantage with respect to the continuous algorithms.
    
    For linUCB the $\alpha$ parameter is set to 0.05, which was chosen from a hyperparameter tuning of values between 0 and 10, while for linTS the $v$ parameter is fixed at 1 for the whole simulation.

    One difficulty with the neural network policies is that they run into the cold-start problem, where the they are unable to make any decisions before they have been trained on some data. For these methods, we set up a warm-up phase in which the policy randomly explores the action space by sampling the actions from a uniform distribution. After this initial phase, the algorithm should have enough data to train. In the experiments we set a warm-up time of 1000 steps, with actions sampled from a uniform distribution $a\sim U(0.5,5.5)$.
    
    Both the \emph{NN Discrete} and the \emph{NN HMC} algorithms share the same neural network architecture, which is a simple multilayer perceptron (MLP) with two hidden layers of 50 neurons with ReLu activation functions and a single neuron with a sigmoid activation function on the output. They are trained after every 100 steps for 10 epochs with a batch size of 2. The alpha entropy exploration term is set to 0.1 and 0.5 for \textit{NN discrete} and the \textit{NN HMC} respectively.
    
    For the  \textit{NN HMC} algorithm, the initial state of the MCMC is set to 2.5, with a step size of 1 and 100 burn in steps. Adam \citep{kingma2014adam} is used as an optimiser over the binary cross-entropy loss with a learning rate of 0.001.
    
    For the EBM algorithm, both $f_\phi$ and $g_\psi$ share a the same architecture with sightly different parameters. They are both MLPs composed of 4 layers, each of which have sigmoid activation functions, except the last block which is just a linear output. Sigmoids were chosen as they were shown empirically to perform better than ReLu activation functions.
    
    The first two block of $f_\phi$ have an output size of 256, while the following two 128 and 1 respectively. Instead, $g_\psi$ is composed of internal layers with an output size of 128.
    
    The EBM is trained every 100 environment steps for a total of 150 epochs, with a learning rate of 0.005, a dropout of 0.2 on every internal layer, and a batch size of 128. For each training iteration after the first, the weights of the MLPs are initialised as the resulting weights of the previous training run. For the action sampling, the exploration term $\alpha$ is set to 10, the gradient step size $\eta$ is set to 0.2, $\omega$ is sampled from a gaussian with $\sigma=0.005$ and 100 SGLD steps are carried out to choose the action. 
    
    To deal with instabilities in the training of the EBMs, for the first training iteration we initialise the weights randomly and retrain the model from 10 to 55 times, searching for the model that minimises the difference between a played action and the action that brings a positive reward. Looking at the actions in the warm-up phase that resulted in positive rewards, we evaluate the actions that the model would play, by drawing a single sample from $\pi$. We compute the average absolute difference over all of these samples and take the model that minimises this average, or stop if we find a model with an average difference of less than 0.6.3

\subsection{Results}
\begin{table*}[]
\begin{tabular}{
>{\columncolor[HTML]{FFFFFF}}r 
>{\columncolor[HTML]{FFFFFF}}c 
>{\columncolor[HTML]{FFFFFF}}c 
>{\columncolor[HTML]{FFFFFF}}c 
>{\columncolor[HTML]{FFFFFF}}c 
>{\columncolor[HTML]{FFFFFF}}c 
>{\columncolor[HTML]{FFFFFF}}c 
>{\columncolor[HTML]{FFFFFF}}c 
>{\columncolor[HTML]{FFFFFF}}c }\\\hline\hline

\multicolumn{1}{l}{\cellcolor[HTML]{FFFFFF}}                                                & 
\multicolumn{4}{c}{\cellcolor[HTML]{FFFFFF}{\color[HTML]{222222} \textbf{LINEAR}}}   &
\multicolumn{4}{c}{\cellcolor[HTML]{FFFFFF}{\color[HTML]{222222} \textbf{CIRCLE}}}   \\

\multicolumn{1}{l}{\cellcolor[HTML]{FFFFFF}}                                                & 
\multicolumn{2}{c}{\cellcolor[HTML]{FFFFFF}{\color[HTML]{222222} \textbf{STATIC}}}   &
\multicolumn{2}{c}{\cellcolor[HTML]{FFFFFF}{\color[HTML]{222222} \textbf{DYNAMIC}}}  &
\multicolumn{2}{c}{\cellcolor[HTML]{FFFFFF}{\color[HTML]{222222} \textbf{STATIC}}}   &
\multicolumn{2}{c}{\cellcolor[HTML]{FFFFFF}{\color[HTML]{222222} \textbf{DYNAMIC}}}  

\\\hline
\multicolumn{1}{c}{\cellcolor[HTML]{FFFFFF}\textbf{Experiment}} & \textbf{\begin{tabular}[c]{@{}c@{}}Average\\ stochastic\\ regret\end{tabular}} & \textbf{\begin{tabular}[c]{@{}c@{}}Best\\ stocastic\\ regret\end{tabular}} & \textbf{\begin{tabular}[c]{@{}c@{}}Average\\ stochastic\\ regret\end{tabular}} & \textbf{\begin{tabular}[c]{@{}c@{}}Best\\ stocastic\\ regret\end{tabular}} & \textbf{\begin{tabular}[c]{@{}c@{}}Average\\ stochastic\\ regret\end{tabular}} & \textbf{\begin{tabular}[c]{@{}c@{}}Best\\ stocastic\\ regret\end{tabular}} & \textbf{\begin{tabular}[c]{@{}c@{}}Average\\ stochastic\\ regret\end{tabular}} & \textbf{\begin{tabular}[c]{@{}c@{}}Best\\ stocastic\\ regret\end{tabular}} \\\hline
EBM         & \textbf{551 ± 123}    & \textbf{450}  & \textbf{3275 ± 59}    & \textit{3194}     & \textbf{987 ± 402}    & \textbf{613}  & \textbf{3307 ± 71}    & \textbf{3208}\\
NN HMC      & 2229 ± 100            & 2080          & 3578 ± 88             & 3502              & 2074 ± 254            & 1642          & \textit{3572 ± 42}    & \textit{3532} \\
NN Discrete & 1166 ± 98             & 1046          & 3720 ± 75             & 3627              & \textit{991 ± 66}     & \textit{879}  & 3645 ± 37             & 3616 \\
UCB1        & 3721 ± 66             & 3675          & 4508 ± 127            & 4319              & 3683 ± 68             & 3608          & 4002 ± 63             & 3900 \\
TS          & 3552 ± 87             & 3480          & 4868 ± 65             & 4808              & 3619 ± 24             & 3583          & 4760 ± 65             & 4656 \\
linUCB      & 1196 ± 632            & 633           & 3822 ± 497            & \textbf{3134}     & 4446 ± 558            & 3955          & 4933 ± 66             & 4842 \\
linTS	    & \textit{558 ± 47}       & \textit{481}	& \textit{3447 ± 50}    & 3375	            & 4473±56	            & 4413	        & 5012 ± 53	            & 4952 \\
%

\hline
\hline
\end{tabular}
\caption{Average stochastic regret and best stochastic regret of the considered algorithms over five runs for different environments: non-linearly separable (Circle) and linearly separable (Linear) both static and dynamic.} 
\label{tab:results}
\end{table*}


In Table~\ref{tab:results} we report the results of all the considered algorithms over all the aforementioned environments: linearly separable (Linear) and non-linearly separable (Circle), both static and dynamic. In each case the regret is calculated as shown in Eq.~\ref{eq:regret}, where the best possible reward is obtained from the simulation environment. Each run is carried out 5 times, with the mean and standard deviation of stochastic regret reported in the table. 
    
Across most of the experiments the EBM algorithm performed the best, showing a good ability to adapt to both linear and non-linear contexts, along with some adaptability to dynamic environments. The main difficulties with the EBM algorithm came from instabilities in training, which could occasionally lead to a bad performance. This resulted in the large standard deviation in the regret for the static Circle environment.

The \emph{NN Discrete} algorithm also performed well, especially with non-linear contexts, but had less flexibility to deal with dynamic environments. In these cases, the ability to carry out continuous action sampling brought an advantage to \emph{NN HMC}. However, the continuous action sampling was less competitive in static environments. This could likely be improved by further tuning the sample step size, reducing it in environments with less variability.

The UCB1 and TS algorithms, which don't take the context into account, were not able to compete with the algorithms that did. However, they did provide a useful baseline for tuning the other algorithms. The linUCB and the linTS algorithms, which do take the contexts into account, were competitive in the linearly separable environments, but couldn't deal with the non-linearly separable environment.

\section{CONCLUSION}

We have introduced algorithms to solve contextual bandits in both continuous and discrete action spaces, making use of maximum entropy exploration. These algorithms are based on neural networks and work either by estimating the reward given a particular action and context, or by modelling the best action probability with an energy function.

Overall, the EBM algorithm performed best in a series of simulation experiments, showing good potential for applications to contextual bandit problems with continuous action spaces. In discrete action spaces, the \emph{NN Discrete} algorithm also performed comparably well in the non-linear case and suffered from fewer training instabilities.

In future work it would be useful to research other techniques to reduce the instability of training the EBM. It would also be worth investigating more intelligent cold-start policies to improve algorithm performances in the initial steps.

\bibliographystyle{plainnat}  
\bibliography{ref}

\begin{thebibliography}{46}
\providecommand{\natexlab}[1]{#1}
\providecommand{\url}[1]{\texttt{#1}}
\expandafter\ifx\csname urlstyle\endcsname\relax
  \providecommand{\doi}[1]{doi: #1}\else
  \providecommand{\doi}{doi: \begingroup \urlstyle{rm}\Url}\fi

\bibitem[Abbasi-Yadkori et~al.(2011)Abbasi-Yadkori, P{\'a}l, and
  Szepesv{\'a}ri]{abbasi2011improved}
Yasin Abbasi-Yadkori, D{\'a}vid P{\'a}l, and Csaba Szepesv{\'a}ri.
\newblock Improved algorithms for linear stochastic bandits.
\newblock \emph{Advances in neural information processing systems}, 24, 2011.

\bibitem[Adams et~al.(2013)Adams, Shipp, and Friston]{Adams2013}
Rick~A. Adams, Stewart Shipp, and Karl~J. Friston.
\newblock Predictions not commands: active inference in the motor system.
\newblock \emph{Brain Structure and Function}, 218\penalty0 (3):\penalty0
  611--643, 2013.
\newblock \doi{10.1007/s00429-012-0475-5}.
\newblock URL \url{https://doi.org/10.1007/s00429-012-0475-5}.

\bibitem[Agrawal and Goyal(2013)]{agrawal2013thompson}
Shipra Agrawal and Navin Goyal.
\newblock Thompson sampling for contextual bandits with linear payoffs.
\newblock In \emph{International conference on machine learning}, pages
  127--135. PMLR, 2013.

\bibitem[Betancourt and Girolami(2015)]{betancourt2015hamiltonian}
Michael Betancourt and Mark Girolami.
\newblock Hamiltonian monte carlo for hierarchical models.
\newblock \emph{Current trends in Bayesian methodology with applications},
  79\penalty0 (30):\penalty0 2--4, 2015.

\bibitem[Bietti et~al.(2021)Bietti, Agarwal, and
  Langford]{bietti2021contextual}
Alberto Bietti, Alekh Agarwal, and John Langford.
\newblock A contextual bandit bake-off.
\newblock \emph{J. Mach. Learn. Res.}, 22:\penalty0 133--1, 2021.

\bibitem[Boney et~al.(2020)Boney, Kannala, and Ilin]{boney2020regularizing}
Rinu Boney, Juho Kannala, and Alexander Ilin.
\newblock Regularizing model-based planning with energy-based models.
\newblock In \emph{Conference on Robot Learning}, pages 182--191. PMLR, 2020.

\bibitem[Bouneffouf et~al.(2020)Bouneffouf, Rish, and
  Aggarwal]{bouneffouf2020survey}
Djallel Bouneffouf, Irina Rish, and Charu Aggarwal.
\newblock Survey on applications of multi-armed and contextual bandits.
\newblock In \emph{2020 IEEE Congress on Evolutionary Computation (CEC)}, pages
  1--8. IEEE, 2020.

\bibitem[Brown and Friston(2012)]{Brown2012}
Harriet Brown and Karl Friston.
\newblock Free-energy and illusions: The cornsweet effect.
\newblock \emph{Frontiers in Psychology}, 3, 2012.
\newblock ISSN 1664-1078.
\newblock \doi{10.3389/fpsyg.2012.00043}.
\newblock URL
  \url{https://www.frontiersin.org/articles/10.3389/fpsyg.2012.00043}.

\bibitem[Cesa-Bianchi et~al.(2017)Cesa-Bianchi, Gentile, Lugosi, and
  Neu]{cesa2017boltzmann}
Nicol{\`o} Cesa-Bianchi, Claudio Gentile, G{\'a}bor Lugosi, and Gergely Neu.
\newblock Boltzmann exploration done right.
\newblock \emph{Advances in neural information processing systems}, 30, 2017.

\bibitem[Degris et~al.(2012)Degris, Pilarski, and Sutton]{degris2012model}
Thomas Degris, Patrick~M Pilarski, and Richard~S Sutton.
\newblock Model-free reinforcement learning with continuous action in practice.
\newblock In \emph{2012 American Control Conference (ACC)}, pages 2177--2182.
  IEEE, 2012.

\bibitem[Delyon et~al.(1999)Delyon, Lavielle, and
  Moulines]{10.1214/aos/1018031103}
Bernard Delyon, Marc Lavielle, and Eric Moulines.
\newblock {Convergence of a stochastic approximation version of the EM
  algorithm}.
\newblock \emph{The Annals of Statistics}, 27\penalty0 (1):\penalty0 94 -- 128,
  1999.
\newblock \doi{10.1214/aos/1018031103}.
\newblock URL \url{https://doi.org/10.1214/aos/1018031103}.

\bibitem[Dillon et~al.(2017)Dillon, Langmore, Tran, Brevdo, Vasudevan, Moore,
  Patton, Alemi, Hoffman, and Saurous]{dillon2017tensorflow}
Joshua~V Dillon, Ian Langmore, Dustin Tran, Eugene Brevdo, Srinivas Vasudevan,
  Dave Moore, Brian Patton, Alex Alemi, Matt Hoffman, and Rif~A Saurous.
\newblock Tensorflow distributions.
\newblock \emph{arXiv preprint arXiv:1711.10604}, 2017.

\bibitem[Du and Mordatch(2019)]{du2019implicit}
Yilun Du and Igor Mordatch.
\newblock Implicit generation and generalization in energy-based models.
\newblock \emph{arXiv preprint arXiv:1903.08689}, 2019.

\bibitem[Du et~al.(2019{\natexlab{a}})Du, Lin, and Mordatch]{Du2019ModelBP}
Yilun Du, Toru Lin, and Igor Mordatch.
\newblock Model based planning with energy based models.
\newblock \emph{ArXiv}, abs/1909.06878, 2019{\natexlab{a}}.

\bibitem[Du et~al.(2019{\natexlab{b}})Du, Lin, and Mordatch]{Yilun2019}
Yilun Du, Toru Lin, and Igor Mordatch.
\newblock Model based planning with energy based models.
\newblock \emph{CoRR}, abs/1909.06878, 2019{\natexlab{b}}.
\newblock URL \url{http://arxiv.org/abs/1909.06878}.

\bibitem[Friston(2009)]{Friston2009}
Karl Friston.
\newblock The free-energy principle: a rough guide to the brain?
\newblock \emph{Trends in Cognitive Sciences}, 13\penalty0 (7):\penalty0
  293--301, 2022/08/04 2009.
\newblock \doi{10.1016/j.tics.2009.04.005}.
\newblock URL \url{https://doi.org/10.1016/j.tics.2009.04.005}.

\bibitem[Friston(2010)]{Friston2010}
Karl Friston.
\newblock The free-energy principle: a unified brain theory?
\newblock \emph{Nature Reviews Neuroscience}, 11\penalty0 (2):\penalty0
  127--138, 2010.
\newblock \doi{10.1038/nrn2787}.
\newblock URL \url{https://doi.org/10.1038/nrn2787}.

\bibitem[Friston et~al.(2006)Friston, Kilner, and Harrison]{FRISTON200670}
Karl Friston, James Kilner, and Lee Harrison.
\newblock A free energy principle for the brain.
\newblock \emph{Journal of Physiology-Paris}, 100\penalty0 (1):\penalty0
  70--87, 2006.
\newblock ISSN 0928-4257.
\newblock \doi{https://doi.org/10.1016/j.jphysparis.2006.10.001}.
\newblock URL
  \url{https://www.sciencedirect.com/science/article/pii/S092842570600060X}.
\newblock Theoretical and Computational Neuroscience: Understanding Brain
  Functions.

\bibitem[Gopalan et~al.(2014)Gopalan, Mannor, and Mansour]{pmlr-v32-gopalan14}
Aditya Gopalan, Shie Mannor, and Yishay Mansour.
\newblock Thompson sampling for complex online problems.
\newblock In Eric~P. Xing and Tony Jebara, editors, \emph{Proceedings of the
  31st International Conference on Machine Learning}, volume~32 of
  \emph{Proceedings of Machine Learning Research}, pages 100--108, Bejing,
  China, 22--24 Jun 2014. PMLR.
\newblock URL \url{https://proceedings.mlr.press/v32/gopalan14.html}.

\bibitem[Grathwohl et~al.(2019)Grathwohl, Wang, Jacobsen, Duvenaud, Norouzi,
  and Swersky]{grathwohl2019your}
Will Grathwohl, Kuan-Chieh Wang, J{\"o}rn-Henrik Jacobsen, David Duvenaud,
  Mohammad Norouzi, and Kevin Swersky.
\newblock Your classifier is secretly an energy based model and you should
  treat it like one.
\newblock \emph{arXiv preprint arXiv:1912.03263}, 2019.

\bibitem[Haarnoja et~al.(2017)Haarnoja, Tang, Abbeel, and
  Levine]{haarnoja2017reinforcement}
Tuomas Haarnoja, Haoran Tang, Pieter Abbeel, and Sergey Levine.
\newblock Reinforcement learning with deep energy-based policies.
\newblock In \emph{International conference on machine learning}, pages
  1352--1361. PMLR, 2017.

\bibitem[Heess et~al.(2013)Heess, Silver, and Teh]{pmlr-v24-heess12a}
Nicolas Heess, David Silver, and Yee~Whye Teh.
\newblock Actor-critic reinforcement learning with energy-based policies.
\newblock In Marc~Peter Deisenroth, Csaba Szepesvári, and Jan Peters, editors,
  \emph{Proceedings of the Tenth European Workshop on Reinforcement Learning},
  volume~24 of \emph{Proceedings of Machine Learning Research}, pages 45--58,
  Edinburgh, Scotland, 30 Jun--01 Jul 2013. PMLR.
\newblock URL \url{https://proceedings.mlr.press/v24/heess12a.html}.

\bibitem[Kaelbling et~al.(1996)Kaelbling, Littman, and
  Moore]{kaelbling1996reinforcement}
Leslie~Pack Kaelbling, Michael~L Littman, and Andrew~W Moore.
\newblock Reinforcement learning: A survey.
\newblock \emph{Journal of artificial intelligence research}, 4:\penalty0
  237--285, 1996.

\bibitem[Kaiser et~al.(2019)Kaiser, Babaeizadeh, Milos, Osinski, Campbell,
  Czechowski, Erhan, Finn, Kozakowski, Levine, et~al.]{kaiser2019model}
Lukasz Kaiser, Mohammad Babaeizadeh, Piotr Milos, Blazej Osinski, Roy~H
  Campbell, Konrad Czechowski, Dumitru Erhan, Chelsea Finn, Piotr Kozakowski,
  Sergey Levine, et~al.
\newblock Model-based reinforcement learning for atari.
\newblock \emph{arXiv preprint arXiv:1903.00374}, 2019.

\bibitem[Kassraie and Krause(2022)]{kassraie2022neural}
Parnian Kassraie and Andreas Krause.
\newblock Neural contextual bandits without regret.
\newblock In \emph{International Conference on Artificial Intelligence and
  Statistics}, pages 240--278. PMLR, 2022.

\bibitem[Kingma and Ba(2014)]{kingma2014adam}
Diederik~P Kingma and Jimmy Ba.
\newblock Adam: A method for stochastic optimization.
\newblock \emph{arXiv preprint arXiv:1412.6980}, 2014.

\bibitem[Kuleshov and Precup(2014)]{kuleshov2014algorithms}
Volodymyr Kuleshov and Doina Precup.
\newblock Algorithms for multi-armed bandit problems.
\newblock \emph{arXiv preprint arXiv:1402.6028}, 2014.

\bibitem[Lai and Robbins(1985)]{lai1985asymptotically}
Tze~Leung Lai and Herbert Robbins.
\newblock Asymptotically efficient adaptive allocation rules.
\newblock \emph{Advances in applied mathematics}, 6\penalty0 (1):\penalty0
  4--22, 1985.

\bibitem[LeCun et~al.(2006)LeCun, Chopra, Hadsell, Ranzato, and
  Huang]{lecun2006tutorial}
Yann LeCun, Sumit Chopra, Raia Hadsell, M~Ranzato, and F~Huang.
\newblock A tutorial on energy-based learning.
\newblock \emph{Predicting structured data}, 1\penalty0 (0), 2006.

\bibitem[Lee et~al.(2020)Lee, Choy, Choi, Kee, and Oh]{lee2020no}
Kyungjae Lee, Jaegu Choy, Yunho Choi, Hogun Kee, and Songhwai Oh.
\newblock No-regret shannon entropy regularized neural contextual bandit online
  learning for robotic grasping.
\newblock In \emph{2020 IEEE/RSJ International Conference on Intelligent Robots
  and Systems (IROS)}, pages 9620--9625. IEEE, 2020.

\bibitem[Levine(2018)]{levine2018reinforcement}
Sergey Levine.
\newblock Reinforcement learning and control as probabilistic inference:
  Tutorial and review.
\newblock \emph{arXiv preprint arXiv:1805.00909}, 2018.

\bibitem[Lippe()]{debmtut}
Phillip Lippe.
\newblock Tutorial 8: Deep energy-based generative models.
\newblock
  \url{https://uvadlc-notebooks.readthedocs.io/en/latest/tutorial_notebooks/tutorial8/Deep_Energy_Models.html}.
\newblock \\Accessed: 2022-07-22.

\bibitem[Markovi{\'c} et~al.(2021)Markovi{\'c}, Stoji{\'c}, Schw{\"o}bel, and
  Kiebel]{markovic2021empirical}
Dimitrije Markovi{\'c}, Hrvoje Stoji{\'c}, Sarah Schw{\"o}bel, and Stefan~J
  Kiebel.
\newblock An empirical evaluation of active inference in multi-armed bandits.
\newblock \emph{Neural Networks}, 144:\penalty0 229--246, 2021.

\bibitem[Moerland et~al.(2020)Moerland, Broekens, and
  Jonker]{moerland2020model}
Thomas~M Moerland, Joost Broekens, and Catholijn~M Jonker.
\newblock Model-based reinforcement learning: A survey.
\newblock \emph{arXiv preprint arXiv:2006.16712}, 2020.

\bibitem[Neal et~al.(2011)]{neal2011mcmc}
Radford~M Neal et~al.
\newblock Mcmc using hamiltonian dynamics.
\newblock \emph{Handbook of markov chain monte carlo}, 2\penalty0
  (11):\penalty0 2, 2011.

\bibitem[Portugal et~al.(2018)Portugal, Alencar, and Cowan]{portugal2018use}
Ivens Portugal, Paulo Alencar, and Donald Cowan.
\newblock The use of machine learning algorithms in recommender systems: A
  systematic review.
\newblock \emph{Expert Systems with Applications}, 97:\penalty0 205--227, 2018.

\bibitem[Riquelme et~al.(2018)Riquelme, Tucker, and Snoek]{riquelme2018deep}
Carlos Riquelme, George Tucker, and Jasper Snoek.
\newblock Deep bayesian bandits showdown: An empirical comparison of bayesian
  deep networks for thompson sampling.
\newblock \emph{arXiv preprint arXiv:1802.09127}, 2018.

\bibitem[Sarker(2021)]{Sarker}
Iqbal~H. Sarker.
\newblock Machine learning: Algorithms, real-world applications and research
  directions.
\newblock \emph{SN Computer Science}, 2\penalty0 (3):\penalty0 160, 2021.
\newblock \doi{10.1007/s42979-021-00592-x}.
\newblock URL \url{https://doi.org/10.1007/s42979-021-00592-x}.

\bibitem[Schwartenbeck et~al.(2013)Schwartenbeck, FitzGerald, Dolan, and
  Friston]{Schwartenbeck2013}
Philipp Schwartenbeck, Thomas FitzGerald, Ray Dolan, and Karl Friston.
\newblock Exploration, novelty, surprise, and free energy minimization.
\newblock \emph{Frontiers in Psychology}, 4, 2013.
\newblock ISSN 1664-1078.
\newblock \doi{10.3389/fpsyg.2013.00710}.
\newblock URL
  \url{https://www.frontiersin.org/articles/10.3389/fpsyg.2013.00710}.

\bibitem[Silver et~al.(2016)Silver, Huang, Maddison, Guez, Sifre, van~den
  Driessche, Schrittwieser, Antonoglou, Panneershelvam, Lanctot, Dieleman,
  Grewe, Nham, Kalchbrenner, Sutskever, Lillicrap, Leach, Kavukcuoglu, Graepel,
  and Hassabis]{Silver}
David Silver, Aja Huang, Chris~J. Maddison, Arthur Guez, Laurent Sifre, George
  van~den Driessche, Julian Schrittwieser, Ioannis Antonoglou, Veda
  Panneershelvam, Marc Lanctot, Sander Dieleman, Dominik Grewe, John Nham, Nal
  Kalchbrenner, Ilya Sutskever, Timothy Lillicrap, Madeleine Leach, Koray
  Kavukcuoglu, Thore Graepel, and Demis Hassabis.
\newblock Mastering the game of go with deep neural networks and tree search.
\newblock \emph{Nature}, 529\penalty0 (7587):\penalty0 484--489, 2016.
\newblock \doi{10.1038/nature16961}.
\newblock URL \url{https://doi.org/10.1038/nature16961}.

\bibitem[Smith et~al.(2022)Smith, Friston, and Whyte]{smith2022step}
Ryan Smith, Karl~J Friston, and Christopher~J Whyte.
\newblock A step-by-step tutorial on active inference and its application to
  empirical data.
\newblock \emph{Journal of mathematical psychology}, 107:\penalty0 102632,
  2022.

\bibitem[Song and Ermon(2019)]{song2019generative}
Yang Song and Stefano Ermon.
\newblock Generative modeling by estimating gradients of the data distribution.
\newblock \emph{Advances in Neural Information Processing Systems}, 32, 2019.

\bibitem[Sutton and Barto(2018)]{sutton2018reinforcement}
Richard~S Sutton and Andrew~G Barto.
\newblock \emph{Reinforcement learning: An introduction}.
\newblock MIT press, 2018.

\bibitem[Thompson(1933)]{thompson1933likelihood}
William~R Thompson.
\newblock On the likelihood that one unknown probability exceeds another in
  view of the evidence of two samples.
\newblock \emph{Biometrika}, 25\penalty0 (3-4):\penalty0 285--294, 1933.

\bibitem[Zhang et~al.(2020)Zhang, Zhou, Li, and Gu]{zhang2020neural}
Weitong Zhang, Dongruo Zhou, Lihong Li, and Quanquan Gu.
\newblock Neural thompson sampling.
\newblock \emph{arXiv preprint arXiv:2010.00827}, 2020.

\bibitem[Zhou et~al.(2020)Zhou, Li, and Gu]{zhou2020neural}
Dongruo Zhou, Lihong Li, and Quanquan Gu.
\newblock Neural contextual bandits with ucb-based exploration.
\newblock In \emph{International Conference on Machine Learning}, pages
  11492--11502. PMLR, 2020.

\end{thebibliography}

\end{document}